\documentclass[3p,times,procedia]{elsarticle}
\usepackage{ecrc}
\usepackage[bookmarks=false]{hyperref}
    \hypersetup{colorlinks,
      linkcolor=blue,
      citecolor=blue,
      urlcolor=blue}

\usepackage{amssymb}

\usepackage[figuresright]{rotating}
\begin{document}
\begin{frontmatter}


\title{Soccer captioning: dataset, transformer-based model, and triple-level evaluation}

\author[a,b]{Ahmad Hammoudeh\corref{cor1}} 
\author[b,c]{Bastien Vanderplaetse}
\author[b]{Stéphane Dupont}

\address[a]{ISIA Lab, Belgium}
\address[b]{MAIA Lab, Belgium}
\address[c]{MARO Lab, Belgium}
\begin{abstract}
This work aims at generating captions for soccer videos using deep learning. The paper introduces a novel dataset, model, and triple-level evaluation. The dataset consists of 22k caption-clip pairs and three visual features (images, optical flow, inpainting) for ~500 hours of \emph{SoccerNet} videos. The model is divided into three parts: a transformer learns language, ConvNets learn vision, and a fusion of linguistic and visual features generates captions. The suggested evaluation criterion of captioning models covers three levels: syntax (the commonly used evaluation metrics such as BLEU-score and CIDEr), semantics (the quality of descriptions for a domain expert), and corpus (the diversity of generated captions). The paper shows that the diversity of generated captions has improved (from 0.07 reaching 0.18) with  semantics-related losses that prioritize selected words. Semantics-related losses and the utilization of more visual features (optical flow, inpainting) improved the normalized captioning score by 27\%.
\end{abstract}

\cortext[cor1]{Corresponding author.}
\end{frontmatter}
\email{A}

\graphicspath{ ./images/ }


\section{Introduction}
\label{main}

\begin{figure}[h]
\includegraphics[scale=0.55]{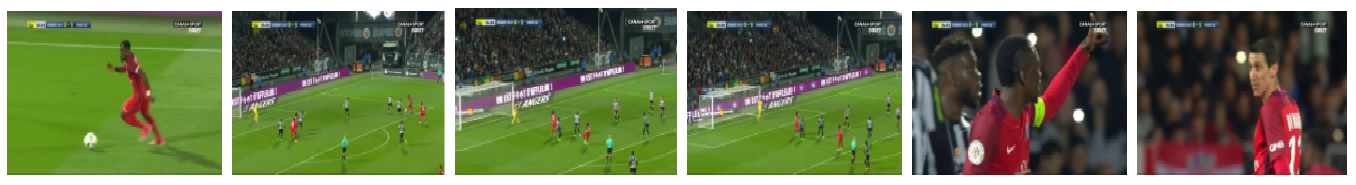}
  \caption{Generated caption: \{player\} \{team\} unleashes a shot, but his effort is poor and floats high over the bar.\\
Ground truth: \{player\} \{team\} unleashes a shot towards the goal, but his effort is not precise at all and it flies high over the bar.}
  \label{fig:fig1}
\end{figure}

Before the spread of televisions, sports fans followed games by listening to sports commentators on the radio \cite{crisell2005introductory}. Since then, commentary has been an essential part of sports broadcasting. A sports commentator tells what is happening in a game and highlights main events such as goals and substitutions. Soccer commentators are typically humans with sports knowledge. That knowledge comes in two folds: A) a low level of knowledge extracted from what is happening inside the pitch (e.g., goal, pass, and penalty), B) and a higher level of knowledge that relies on external information such as the context of a game, and the history of teams /players. The higher level of knowledge cannot be extracted by just watching a match. For example, when a commentator says: “\{player\} scores his seventh goal in the tournament,” the commentator relies on a statistic that the player scored six other goals in the tournament.\\ \\
Creativity is an additional dimension of soccer commentary. Some soccer commentators use figurative language or rhyming sentences to describe important actions. For example, a commentator described a goal in the Spanish league as firing a ball past a diving goalkeeper. Another commentator, I.Chawali, described a player as Charlie Chaplin presenting a silent show, and Alexander Fleming treating his team with Penicillin.” Although those  statements do not provide facts about the game, they entertain listeners. In \cite{schultz2012sports}, Schultz elaborated on the profound factor of entertainment in the sports broadcasting industry. Sports commentary is not just reporting a game, but it is also entertaining, metaphorical, and emotional. Deep learning could be a potential approach toward machine-based soccer commentary systems that imitate human commentators. This work introduces the first soccer captioning system (dataset, deep learning model, and multi-level evaluation criterion). Fig. \ref{fig:fig1} shows an example of a generated cation. More examples with videos are available on the web page of this paper:' https://sites.google.com/view/soccercaptioning '.

\section{Related work}
\label{sec:headings}

\textbf{Generic video captioning}: Earlier approaches to converting videos into sentences relied on template-based hierarchical language models \cite{guadarrama2013youtube2text, rohrbach2013translating}. The idea was to identify a set of categories in a video and predict semantic relations between them to generate a sentence according to a pre-defined ontology (i.e., a person does something). Later, deep learning replaced template-based approaches. The encoder-decoder architectures \cite{venugopalan2014translating} transform content from a visual domain(video) into a linguistic domain (text) through an intermediate domain (hidden state). The encoder transforms the visual features into a hidden state, and the decoder transforms the hidden state into a linguistic representation. After the attention mechanism’s breakthrough in the area of natural language processing, it was implemented in video captioning \cite{yao2015describing, zhou2019grounded, pan2020spatio}. Multimodality is known to improve the outcome of machine learning models \cite{delbrouck2018umons, delbrouck2017modulating, delbrouck2017empirical, delbrouck2017visually, delbrouck2017multimodal}. Video captioning typically relied on visual features only extracted from video frames such as Resnet and I3D features. However, including more modalities like audio and speech improved video captioning as shown in \cite{hessel2019case, iashin2020multi}.\\ \\
\textbf{Sports video captioning}: Generic video captioning models generate sentences from a macroscopic and general perspective, with no domain-specific information and details. For example, a soccer clip would be captioned as “people play football.” In the basketball domain, a hierarchically grouped recurrent architecture was proposed for more domain-specific and detailed captions  \cite{yu2018fine}. The architecture was a fusion of three parts: 1) a CNN model for pixel-segmentation where every pixel was assigned to one of four categories: ball, first-team, second-team, and background. 2) a model that encodes the movement of individuals using optical-flow features 3) a part for modeling the relationship between players. The three parts were fused in a hierarchically recurrent structure to caption NBA basketball videos. On the same side, attention mechanisms with hierarchical recurrent neural networks were used to caption volleyball videos \cite{qi2019sports}. In the context of captioning soccer games using deep learning, no similar work was reported to the best of the authors’ knowledge. However, humanoid commentators were developed for soccer games played by robotic dogs \cite{veloso2008team}. The humanoid commentators selected an utterance from a predefined library using a rule-based event identification. The event identification depended on the game history and event recognition using a SIFT-based vision analysis algorithm.\\ \\
\textbf{SoccerNet} is a large-scale dataset of 500 soccer games \cite{deliege2021soccernet}. It was used for developing soccer action spotting models \cite{tomei2021rms}. Multimodality using both acoustic and visual features improved event detection in general \cite{brousmiche2022multimodal, brousmiche2020intra, brousmiche2021multi, brousmiche2020avecl, brousmiche2019audio} and Soccer in particular. An Improvement of 4.2\% of the mean average precision in soccer action spotting and 7.4\% in action classification were reported \cite{vanderplaetse2020improved}. The captioned actions in this work are based on the actions labeled in SoccerNet.

\section{Proposed methodology}
\label{sec:Proposed method}

This section introduces three pillars needed for developing a soccer captioning system: A dataset, model, and multi-level evaluation criterion. 

\subsection{ Dataset}
The introduced dataset consists of 22k soccer video-caption pairs (statistics are shown in Table \ref{table:table1}). The videos were taken from SoccerNet dataset. The captions were crawled from flashscore.com website. Captions were formatted. The names of players, coaches, and teams were replaced by single tokens representing each name's category (player, coach, team, time). For example, ‘Christiano Ronolado’ was replaced by ‘\{player\}’ and ‘José Mourinho’ was replaced by ‘\{coach\}’. The Identification of the individual names entails external information, such as mapping between players' names and their numbers/faces, that cannot be learned from the created dataset.

\begin{table}[h]
    \centering
\begin{center}
\caption{Soccer captioning dataset - number of captions per action}
\begin{tabular}{ c c c c  c c c  c c c  c c c  c c c  c c c }
\hline
\rotatebox[origin=c]{90}{shots on target}  
&\rotatebox[origin=c]{90}{corner}  
&\rotatebox[origin=c]{90}{substitution}  
&\rotatebox[origin=c]{90}{yellowcard}  
&\rotatebox[origin=c]{90}{Shots off target}
&\rotatebox[origin=c]{90}{foul} 
&\rotatebox[origin=c]{90}{kick-off}  
&\rotatebox[origin=c]{90}{ball out of play}  
&\rotatebox[origin=c]{90}{goal}  
&\rotatebox[origin=c]{90}{direct freekick}  
&\rotatebox[origin=c]{90}{offside}  
&\rotatebox[origin=c]{90}{ indirect freekick }  &\rotatebox[origin=c]{90}{penalty}  
&\rotatebox[origin=c]{90}{redcard}  
&\rotatebox[origin=c]{90}{clearance}  
&\rotatebox[origin=c]{90}{yellow-red card} \\ 
\hline
2326 & 3891 & 2171 & 1646 & 2528 & 3085 &  769 & 1252&
1295 & 795 & 1267 & 399 & 87 & 44 & 68  & 33\\

\hline
\end{tabular}

\label{table:table1}
\end{center}
\end{table}

\subsubsection{ Multiple features}\label{subsection:Features}
\begin{figure}[h]
\centerline{\includegraphics[width=0.15\linewidth, height=1.2cm]{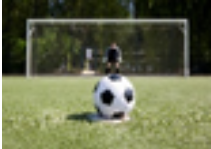}\hspace*{15mm}\includegraphics[width=0.3\linewidth, height=2cm]{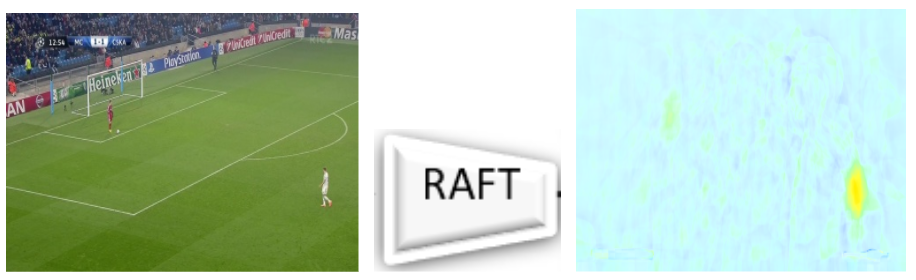}
\hspace*{15mm}\includegraphics[width=0.3\linewidth, height=2cm]{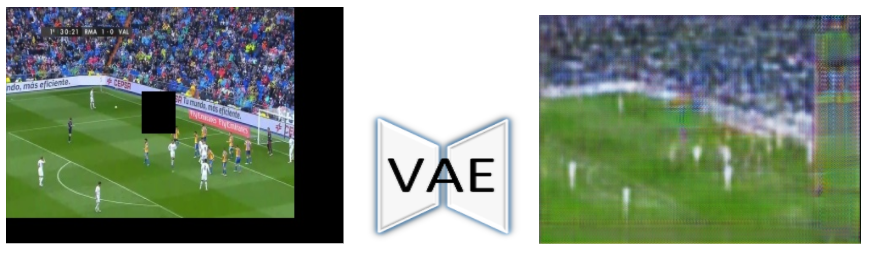}
}
\caption{ Visual features: (a) RGB image; (b) Optical flow features extracted using the RAFT model. The moving players are highlighted in yellow; (c) Image inpainting. From left to right: input image with missing parts, VAE, and reconstructed image.}
\label{fig:image2}
\end{figure}

Learning from videos is challenging due to the problem of high dimensionality. Videos were represented using low-dimensional features to overcome the heavy processing of spatio-temporal data. Extracting multiple features provides more clues about the video. Three features (see Fig. \ref{fig:image2}) were extracted: images, optical flow, and image inpainting features.

\begin{itemize}
\item \textbf{RGB images (img)} extracted at a frame rate of 1 image every half a second (2 FPS). The resolution was reduced to 32x64x3 pixels to speed up the computation. The intuition behind the low-resolution images is to provide clues about the relative locations of visible objects in a frame.
\item \textbf{Optical flow features (flow)} were extracted using a pre-trained optical-flow transformer (RAFT: Recurrent all-pairs field transforms \cite{teed2020raft}). The intuition behind optical-flow features is to provide clues about the displacement of objects (players, ball, referees). For more efficient computation, optical flow features were reduced from 2x398x224 to 2x256 using Principal Components Analysis (PCA). A PCA matrix was calculated for a single match and  applied to other matches achieving a retained variance of 98\%.
\item \textbf{Image inpainting features (vae)}. The intuition of extracting features from an image inpainting model is to substitute some missing clues that may not be available in the low-resolution images or the optical flow features. The task of image inpainting is the process of completing the missing parts of an image. A variational auto-encoder (VAE) was trained to reconstruct missing parts of an image achieving a reconstruction loss of 0.08. The missing parts are a horizontal rectangle of 398x24 pixels on the bottom, a vertical rectangle of 224x48 pixels on the right, and a square of 40x40 pixels at the center. The inpainting features were extracted from the latent space of VAE. The VAE was trained on 1 million soccer images.
\end{itemize}

\subsection{ The model : deep soccer captioning }

The proposed model for captioning soccer actions is a word sequence model. The model predicts the next word given a soccer clip and a sequence of previous words. The architecture is a fusion of a transformer and ConvNets. The main differences between the proposed model here and the transformer as in the attention paper \cite{vaswani2017attention} are: 1) that the encoder in this work is a set of ConvNets instead of attention, and 2) a set of fully connected layers replaces the cross attention. The model consists of 3 main parts shown in Fig. \ref{fig:fig5}:

\begin{figure}[b]
\centering
\includegraphics[scale=0.5]{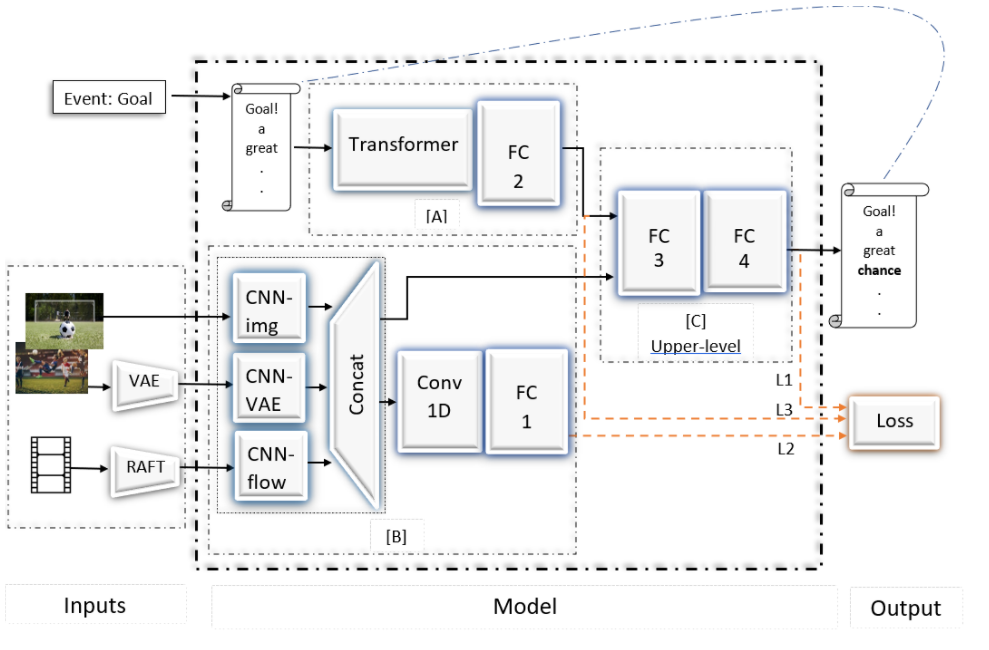}
\caption{ The proposed model}
\label{fig:fig5}
\end{figure}

\begin{enumerate}
\item Part [A] A transformer processes linguistic features and predicts the next word given a sequence of previous words. The transformer is a diminutive version of GPT  \cite{vaswani2017attention} that consists of just a single transformer block and two multi heads while the GPT uses a stack of such blocks.

\item Part [B] consists of ConvNets that process visual features: CNN-img for RGB pixels, CNN-flow for optical flow, and CNN-VAE for inpainting features. The outcomes are then concatenated and passed to part C.

\item Part [C] consists of fully connected layers that take the outputs of part A and part B and yields the final prediction.

\end{enumerate}

\begin{itemize}
\item Tip I: The fully connected network FC2 is followed by two separate heads: one with a sigmoid activation function for the loss and another with Relu activation function that passes the features to FC3. Adding a sigmoid activation layer before sending the loss signal avoids the log(0) error. Relu was selected as the activation function when passing the features to FC3 because ReLU converges faster and avoids the vanishing gradient problem.
\item Tip II: The type of action was fed to the transformer as the first word of the sentence, similar to the Conditional Transformer Language (CTRL), because conditioning captions on the type of action simplifies the captioning task. The state-of-the-art model in SoccerNet-v2 challenge detects soccer actions with a mean average precision of 75\%. 
\end{itemize}

\subsubsection{ Loss functions and prioritized (significant) words }

Words such as goal, free-kick, and penalty indicate specific technical meanings in the soccer game. For example, a goal scored from a penalty is technically different from a goal scored from a free-kick. Moreover, a false generation of a word like ‘penalty’ affects the next sequence of words more than a word such as ‘the’. Learning the significant words (SWs) is prioritized by giving them a greater portion of the overall loss function.\\ \\
Caption generation is learned in a supervised way by feeding three correction signals (losses) at three points of supervision. 
The overall loss is a weighted average of the three losses: The first loss L1 is a categorical cross-entropy loss of the next token fed at the output layer of the captioning model (Part C), as shown in Fig. \ref{fig:fig5}. L2 and L3 focus on the significant words. L3 is a categorical cross-entropy loss of the next token predicted by Part A (the linguistic part). L3 is active for significant words only. L2 is the mean square error of the significant words predicted by Part B (the visual part).\\ \\
Part B assigns 1 to the significant word if it appears in a caption and 0 otherwise. Assume that two significant words appeared in a caption out of ten SWs, the ground truth in this case is a vector of ten digits (the number of SWs), two of them are ones and the rest are zeros. Something like Y\textsubscript{gt} = 0001000010.  The mean square error for all 1s and 0s is MSE( Y\textsubscript{pred} , Y\textsubscript{gt} ), where \emph{Y\textsubscript{pred}} is the prediction of Part B. However, the total number of significant words that are not in the caption (0s) is greater than those in the caption (1s). If a model predicts 0 always, the prediction will be correct for 80\% of the digits. To overcome this, a term was added to the loss which is the mean square error of the ones only: MSE( Y\textsubscript{pred} x Y\textsubscript{gt}, Y\textsubscript{gt} ). hence, L2 is defined as
\begin{equation}\label{equation:eq7}
   L2 = MSE( Y_{pred} , Y_{gt} )  + sc\times MSE( Y_{pred}\times Y_{gt}, Y_{gt} )
\end{equation}

The second part (loss of 1s) was multiplied by a scaling factor (sc) to balance the loss of the 1s and the loss of 0s. Ideally, it is the ratio between the average number of 1’s and 0s. For the soccer captioning dataset, the scaling factor is 20. The semantic loss signal L2 was taken from a head added to the visual part. The head is a convolution layer and a fully connected layer added on top of the visual part. The fully connected layer predicts the significant words in the clip and L2 penalizes wrong predictions.

\subsection{ Evaluation }

\begin{figure}[h]
\centering
\includegraphics[scale=0.4]{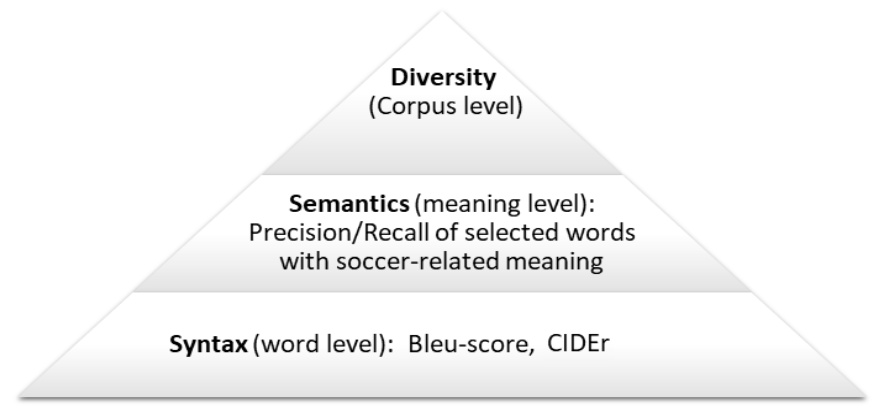}
\caption{ Three levels for evaluating soccer captions: Syntax (word level), Soccer-oriented semantics (meaning level), and diversity (corpus level) }
\label{fig:fig6}
\end{figure}

The evaluation of generated captions included three aspects (see Fig. \ref{fig:fig6}):  Syntax (word level), Soccer-oriented semantics (meaning level), and diversity (corpus level).
\begin{itemize}
\item Syntax-oriented evaluation: Captioning models are typically evaluated by comparing generated captions (hypotheses) against ground truth captions (references), word by word or chunk by chunk. That kind of evaluation is syntax-oriented as any differences at the word level count. This work used BLEU-score and CIDEr from Microsoft CoCo evaluation code \cite{chen2015microsoft}. \\ 
\item Soccer-oriented evaluation (meaning level):
The syntactic similarity between a generated caption and a reference caption does not always reflect the quality of a generated caption in the soccer domain. The example in Table \ref{table:table2} shows almost two identical captions. Although 27 words out of 28 words are similar and one word only is different, that single word makes a technical difference. Such a difference does not appear clearly in syntax-oriented metrics like BLEU score or CIDEr. In order to evaluate the caption from the soccer perspective, another evaluation criterion was proposed: words with no technical meanings are removed, and just the significant words are evaluated using the precision and recall of the stemmed significant words in the caption.

\begin{table}[h]
\caption{A caption demonstrates how the precision of SW detects the technical differences better than generic captioning metrics (Bleu-score)}
\vspace{-8 mm}
\begin{center}
\begin{tabular}{  p{1.5cm}  p{5.4cm}  p{5.4cm}  p{0.9cm}  p{1.3cm}  }
\hline
 & 
Reference & 
Hypothesis & 
B@1 & 
Precision \\
\hline
Caption & 
That was unbelievable. \{PLAYER\} \{TEAM\} changes the scoreline after getting on the end of a brilliant \underline{pass} and firing a precise \underline{shot} that goes inside the \underline{\textbf{right}} \underline{post} & 
That was unbelievable. \{PLAYER\} \{TEAM\} changes the scoreline after getting on the end of a brilliant \underline{pass} and firing a precise \underline{shot} that goes inside the \underline{\textbf{left}} \underline{post} & 
96\% & 
- \\
SWs & 
[ pass, shot, right, post ] & 
[ pass, shot, left, post ] & 
- & 
75\% \\
\hline
\end{tabular}
\label{table:table2}
\end{center}
\end{table}

\item The diversity (corpus level): 
The diversity of generated captions is the ratio of the number of distinct generated captions N\textsubscript{hypo} to the number of distinct ground truth captions N\textsubscript{ref} as in Equation \ref{equation:eq8}.
\begin{equation}\label{equation:eq8}
    Diversity = {N_{hypo} \over N_{ref}}
\end{equation}
\end{itemize}

\section{Experiments and results}
\label{sec:Experimental evaluation}

The dataset was divided randomly into 85\% training, 5\% validation, and 10\% testing. The size of the vocabulary was 1400 tokens after removing words that appeared less than 4 times. Captions were tokenized after converting uppercase letters to lowercase letters. Tokens included words and a set of selected punctuations [ ! . , ]. \\
The list of SWs:  ['goal', 'post', 'pass', 'net', 'corner', 'goalkeeper', 'penalty', 'bar', 'kick', ‘shot’, 'cross','freekick', 'yellow', 'red', 'card', 'area', 'rebound', 'free', 'head', 'offside', 'throw-in', 'box', 'right', 'left','over', 'inside', 'bottom', 'back', 'up', 'side', 'top', ( 'loft', 'float'), 'middle', ' 'outside','high', 'mid-range', 'roof', 'out', 'off', 'first', 'second', 'half', 'long', 'low', 'flag', 'linesman', 'short','defender', 'teammate', 'opponent', ('work', 'decides'), ('replace',  'change','substitution', 'substitute'), ('cut', 'intercept'), ('foul', 'tackle', 'challenge'),('nothing', 'clear', 'save', 'fail', 'waste'','block') ]. The number of SWs is 55 after combining semantically equivalent words (between parentheses).\\ \\
\textbf{The stopping criterion} depends on the captioning metrics, not the loss. The model that yields the best evaluation metrics for the caption generation task is not necessarily that with the best loss. The loss used for word sequence models depends only on one token (the next token). On the other hand, the video captioning metrics evaluate the entire sentence (a  generated sentence against a reference sentence).\\ \\
For the purpose of using a single metric in the stopping criterion, the average of normalized BLEU-score, CIDEr, Precision and Recall was considered. In normalization, each evaluation metric was divided by a nominal value as in Equation \ref{equation:eq9}. The nominal values were the values scored by a baseline model.
\begin{equation}\label{equation:eq9}
    NormalizedScore = {1 \over4} ({{B@4 \over 10} + {CIDEr\over 0.55} + {Precision\over 40} + {Recall\over 40} })
\end{equation}

The proposed model was compared against the following captioning models:
\begin{itemize}
\item baseline: A naive captioning model that generates a random caption given an action category (without visual features). 
\item k-NN: The nearest neighbor approach gives a video the same caption as its closest labeled video (k =1). 
A feature representation model was trained using a triplet loss to estimate the distance between videos based on their captions similarity.
\item Rsnt-L1: a deep captioning model trained on Resnet features using L1 loss only.
\end{itemize}

The detailed results are shown in Table \ref{table:table4}, and the abbreviations are defined below.
\begin{table}[h]
    \centering
\caption{Captioning results and ablation study}
\vspace{-2 mm}
\begin{center}
\begin{tabular}{ p{3.3cm}  p{3.1cm}  
p{3.6 cm}  p{4.6cm} }
\hline
  & 
Syntax & 
Semantics &\\
\end{tabular}
\begin{tabular}{  p{3.3cm}  p{0.7cm}  
p{2cm}  p{1.2 cm}  p{2cm}  p{4.6cm} }
\hline
  & 
B@4 & 
CIDEr& 
Precision & 
Recall & 
Normalized score\\
\hline
Proposed & 
15.1 & 
0.95 & 
49 & 
46.6 & 
1.41\\
w/o L2 & 
13.1 & 
0.84 & 
50.6 & 
51.7 & 
1.35\\
w/o L3 & 
13.2 & 
0.9 & 
49.2 & 
46.4 & 
1.34\\
Rsnt-L1 & 
11.8 & 
0.65 & 
40.6 & 
43.3 & 
1.11\\
k-NN & 
9.8 & 
0.53 & 
42.6 & 
43.6 & 
1.02\\
baseline & 
9.8 & 
0.54 & 
39.8 & 
40 & 
1\\
\hline
\end{tabular}
\label{table:table4}
\end{center}
\end{table}

The proposed approach yielded a normalized score of 1.41. 
The k-NN approach did not surpass the deep learning approach. Moreover, k-NN limits the assigned captions to the available captions (no new sentences). On the other hand, deep learning approaches can create new captions \cite{nikolaus2019compositional}. The semantics-related losses (L3 and L2) boosted the normalized score from 1.35 to 1.41 as shown in Table \ref{table:table4} and boosted the diversity of the generated captions from 0.07 to 0.18.

\section{Conclusion}

This work introduced three main contributions for soccer captioning using deep learning: a dataset, a model, and an evaluation criterion that covers three levels: syntax (the popular evaluation metrics such as BLEU-score and CIDEr), meaning (the quality of description for a domain-expert), and corpus (the diversity of generated captions). It also shows that the diversity of generated captions improved from 0.07 to 0.18 with  semantics-related losses that prioritize selected words. Semantics-related losses along with the utilization of multiple visual features improved the normalized captioning score by 27\%.

\bibliography{references}  

\begin{thebibliography}{10}
\expandafter\ifx\csname url\endcsname\relax
  \def\url#1{\texttt{#1}}\fi
\expandafter\ifx\csname urlprefix\endcsname\relax\def\urlprefix{URL }\fi
\expandafter\ifx\csname href\endcsname\relax
  \def\href#1#2{#2} \def\path#1{#1}\fi

\bibitem{crisell2005introductory}
A.~Crisell, An introductory history of British broadcasting, Routledge, 2005.

\bibitem{schultz2012sports}
B.~Schultz, Sports media: Reporting, producing, and planning, Routledge, 2012.

\bibitem{guadarrama2013youtube2text}
S.~Guadarrama, N.~Krishnamoorthy, G.~Malkarnenkar, S.~Venugopalan, R.~Mooney,
  T.~Darrell, K.~Saenko, Youtube2text: Recognizing and describing arbitrary
  activities using semantic hierarchies and zero-shot recognition, in:
  Proceedings of the IEEE international conference on computer vision, 2013,
  pp. 2712--2719.

\bibitem{rohrbach2013translating}
M.~Rohrbach, W.~Qiu, I.~Titov, S.~Thater, M.~Pinkal, B.~Schiele, Translating
  video content to natural language descriptions, in: Proceedings of the IEEE
  international conference on computer vision, 2013, pp. 433--440.

\bibitem{venugopalan2014translating}
S.~Venugopalan, H.~Xu, J.~Donahue, M.~Rohrbach, R.~Mooney, K.~Saenko,
  Translating videos to natural language using deep recurrent neural networks,
  arXiv preprint arXiv:1412.4729.

\bibitem{yao2015describing}
L.~Yao, A.~Torabi, K.~Cho, N.~Ballas, C.~Pal, H.~Larochelle, A.~Courville,
  Describing videos by exploiting temporal structure, in: Proceedings of the
  IEEE international conference on computer vision, 2015, pp. 4507--4515.

\bibitem{zhou2019grounded}
L.~Zhou, Y.~Kalantidis, X.~Chen, J.~J. Corso, M.~Rohrbach, Grounded video
  description, in: Proceedings of the IEEE/CVF Conference on Computer Vision
  and Pattern Recognition, 2019, pp. 6578--6587.

\bibitem{pan2020spatio}
B.~Pan, H.~Cai, D.-A. Huang, K.-H. Lee, A.~Gaidon, E.~Adeli, J.~C. Niebles,
  Spatio-temporal graph for video captioning with knowledge distillation, in:
  Proceedings of the IEEE/CVF Conference on Computer Vision and Pattern
  Recognition, 2020, pp. 10870--10879.

\bibitem{delbrouck2018umons}
J.-B. Delbrouck, S.~Dupont, Umons submission for wmt18 multimodal translation
  task, arXiv preprint arXiv:1810.06233.

\bibitem{delbrouck2017modulating}
J.-B. Delbrouck, S.~Dupont, Modulating and attending the source image during
  encoding improves multimodal translation, arXiv preprint arXiv:1712.03449.

\bibitem{delbrouck2017empirical}
J.-B. Delbrouck, S.~Dupont, An empirical study on the effectiveness of images
  in multimodal neural machine translation, arXiv preprint arXiv:1707.00995.

\bibitem{delbrouck2017visually}
J.-B. Delbrouck, S.~Dupont, O.~Seddati, Visually grounded word embeddings and
  richer visual features for improving multimodal neural machine translation,
  arXiv preprint arXiv:1707.01009.

\bibitem{delbrouck2017multimodal}
J.-B. Delbrouck, S.~Dupont, Multimodal compact bilinear pooling for multimodal
  neural machine translation, arXiv preprint arXiv:1703.08084.

\bibitem{hessel2019case}
J.~Hessel, B.~Pang, Z.~Zhu, R.~Soricut, A case study on combining asr and
  visual features for generating instructional video captions, arXiv preprint
  arXiv:1910.02930.

\bibitem{iashin2020multi}
V.~Iashin, E.~Rahtu, Multi-modal dense video captioning, in: Proceedings of the
  IEEE/CVF Conference on Computer Vision and Pattern Recognition Workshops,
  2020, pp. 958--959.

\bibitem{yu2018fine}
H.~Yu, S.~Cheng, B.~Ni, M.~Wang, J.~Zhang, X.~Yang, Fine-grained video
  captioning for sports narrative, in: Proceedings of the IEEE Conference on
  Computer Vision and Pattern Recognition, 2018, pp. 6006--6015.

\bibitem{qi2019sports}
M.~Qi, Y.~Wang, A.~Li, J.~Luo, Sports video captioning via attentive motion
  representation and group relationship modeling, IEEE Transactions on Circuits
  and Systems for Video Technology 30~(8) (2019) 2617--2633.

\bibitem{veloso2008team}
M.~Veloso, N.~Armstrong-Crews, S.~Chernova, E.~Crawford, C.~McMillen, M.~Roth,
  D.~Vail, S.~Zickler, A team of humanoid game commentators, International
  Journal of Humanoid Robotics 5~(03) (2008) 457--480.

\bibitem{deliege2021soccernet}
A.~Deliege, A.~Cioppa, S.~Giancola, M.~J. Seikavandi, J.~V. Dueholm,
  K.~Nasrollahi, B.~Ghanem, T.~B. Moeslund, M.~Van~Droogenbroeck, Soccernet-v2:
  A dataset and benchmarks for holistic understanding of broadcast soccer
  videos, in: Proceedings of the IEEE/CVF Conference on Computer Vision and
  Pattern Recognition, 2021, pp. 4508--4519.

\bibitem{tomei2021rms}
M.~Tomei, L.~Baraldi, S.~Calderara, S.~Bronzin, R.~Cucchiara, Rms-net:
  Regression and masking for soccer event spotting, in: 2020 25th International
  Conference on Pattern Recognition (ICPR), IEEE, 2021, pp. 7699--7706.

\bibitem{brousmiche2022multimodal}
M.~Brousmiche, J.~Rouat, S.~Dupont, Multimodal attentive fusion network for
  audio-visual event recognition, Information Fusion 85 (2022) 52--59.

\bibitem{brousmiche2020intra}
M.~Brousmiche, S.~Dupont, J.~Rout, Intra and inter-modality interactions for
  audio-visual event detection, in: Proceedings of the 1st International
  Workshop on Human-centric Multimedia Analysis, 2020, pp. 5--11.

\bibitem{brousmiche2021multi}
M.~Brousmiche, J.~Rouat, S.~Dupont, Multi-level attention fusion network for
  audio-visual event recognition, arXiv preprint arXiv:2106.06736.

\bibitem{brousmiche2020avecl}
M.~Brousmiche, S.~Dupont, J.~Rouat, Avecl-umons database for audio-visual event
  classification and localization, arXiv preprint arXiv:2011.01018.

\bibitem{brousmiche2019audio}
M.~Brousmiche, J.~Rouat, S.~Dupont, Audio-visual fusion and conditioning with
  neural networks for event recognition, in: 2019 IEEE 29th International
  Workshop on Machine Learning for Signal Processing (MLSP), IEEE, 2019, pp.
  1--6.

\bibitem{vanderplaetse2020improved}
B.~Vanderplaetse, S.~Dupont, Improved soccer action spotting using both audio
  and video streams, in: Proceedings of the IEEE/CVF Conference on Computer
  Vision and Pattern Recognition Workshops, 2020, pp. 896--897.

\bibitem{teed2020raft}
Z.~Teed, J.~Deng, Raft: Recurrent all-pairs field transforms for optical flow,
  in: European conference on computer vision, Springer, 2020, pp. 402--419.

\bibitem{vaswani2017attention}
A.~Vaswani, N.~Shazeer, N.~Parmar, J.~Uszkoreit, L.~Jones, A.~N. Gomez,
  {\L}.~Kaiser, I.~Polosukhin, Attention is all you need, in: Advances in
  neural information processing systems, 2017, pp. 5998--6008.

\bibitem{chen2015microsoft}
X.~Chen, H.~Fang, T.-Y. Lin, R.~Vedantam, S.~Gupta, P.~Doll{\'a}r, C.~L.
  Zitnick, Microsoft coco captions: Data collection and evaluation server,
  arXiv preprint arXiv:1504.00325.

\bibitem{nikolaus2019compositional}
M.~Nikolaus, M.~Abdou, M.~Lamm, R.~Aralikatte, D.~Elliott, Compositional
  generalization in image captioning, arXiv preprint arXiv:1909.04402.

\end{thebibliography}
\bibliographystyle{elsarticle-num}

\end{document}